\pdfoutput=1

\documentclass[11pt]{article}

\usepackage[]{acl}
\usepackage{mainlp} 

\usepackage{wrapfig}
\definecolor{light-gray}{gray}{0.95}
\usepackage{tcolorbox}
\usepackage{subcaption}
\usepackage{listings}
\usepackage{color}
\usepackage{soul}
\definecolor{lightgrey}{rgb}{0.925, 0.925, 0.925}
\sethlcolor{lightgrey}
\newcommand{\codebox}[1]{\texttt{\colorbox{lightgray}{#1}}}

\usepackage{todonotes}
\usepackage{booktabs}

\newcommand{\yes}{\ding{51}}
\newcommand{\no}{\ding{53}}
\newcommand{\yesself}{\fbox{\textcolor{magenta}{\ding{51}}}}
\newcommand{\noself}{\fbox{\textcolor{magenta}{\ding{53}}}}
\newcommand{\idk}{\textbf{?}}
\newcommand{\idkself}{\fbox{\textcolor{magenta}{\textbf{?}}}}

\title{\name{} NLI: Separating Annotation Error from Human Label Variation}

\author{Leon Weber-Genzel\textsuperscript{\faMountain}\thanks{\;\; Equal contribution.}
\hspace{2pt}
Siyao Peng\textsuperscript{\faMountain}\footnotemark[1]
\hspace{2pt}
Marie-Catherine de Marneffe\textsuperscript{\faPenFancy}
Barbara Plank\textsuperscript{\faMountain\kern1pt}\\
\textsuperscript{\faMountain} MaiNLP \& MCML, 
LMU Munich, Germany \\
\textsuperscript{\faPenFancy} FNRS, CENTAL, UCLouvain, Belgium \\
{\tt \{siyao.peng,b.plank\}@lmu.de \hspace{2pt}
marie-catherine.demarneffe@uclouvain.be}}

\newcommand{\name}[0]{\textsc{VariErr}}

\begin{document}
\maketitle

\begin{abstract}
Human label variation arises when annotators assign different labels to the same item for valid reasons, while annotation errors occur when labels are assigned for invalid reasons. 
These two issues are prevalent in NLP benchmarks, yet existing research has studied them in isolation. 
To the best of our knowledge, there exists no prior work that focuses on teasing apart error from signal, especially in cases where signal is beyond black-and-white.
To fill this gap, we introduce a systematic methodology and a new dataset, \name{} (variation versus error), focusing on the NLI task in English. We propose a 2-round annotation procedure
with annotators explaining each label and subsequently judging the validity of label-explanation pairs.
\name{} contains 7,732 validity judgments on 1,933 explanations for 500 re-annotated MNLI items. 
We assess the effectiveness of various automatic error detection (AED) methods and GPTs in uncovering errors versus human label variation. We find that state-of-the-art AED methods significantly underperform GPTs and humans. While GPT-4 is the best system, it still falls short of human performance. 
Our methodology is applicable beyond NLI, offering fertile ground for future research on error versus plausible variation, which in turn can yield better and more trustworthy NLP systems.
\end{abstract}

\section{Introduction}\label{sec:introduction}

Labeled data plays a crucial role in modern machine learning (ML)~\cite[e.g.,][]{mazumder2023dataperf}. Data quality impacts ML performance and, in turn, user trust. It is therefore of vital importance to aim at high-quality consistently-labeled benchmark data~\cite[e.g.,][]{bowman-dahl-2021-will}. However, recent research revealed a notable presence of \textit{annotation errors} in widely-used NLP benchmarks~\cite{klie-etal-2023-annotation,rucker-akbik-2023-cleanconll}. Similar observations were made recently in computer vision (CV)~\cite{northcutt2021pervasive,vasudevan2022does,schmarje2023label}.  

\begin{figure}[t]
\centering
\includegraphics[trim=2cm 18cm 34cm 2cm,clip,width=.98\columnwidth]{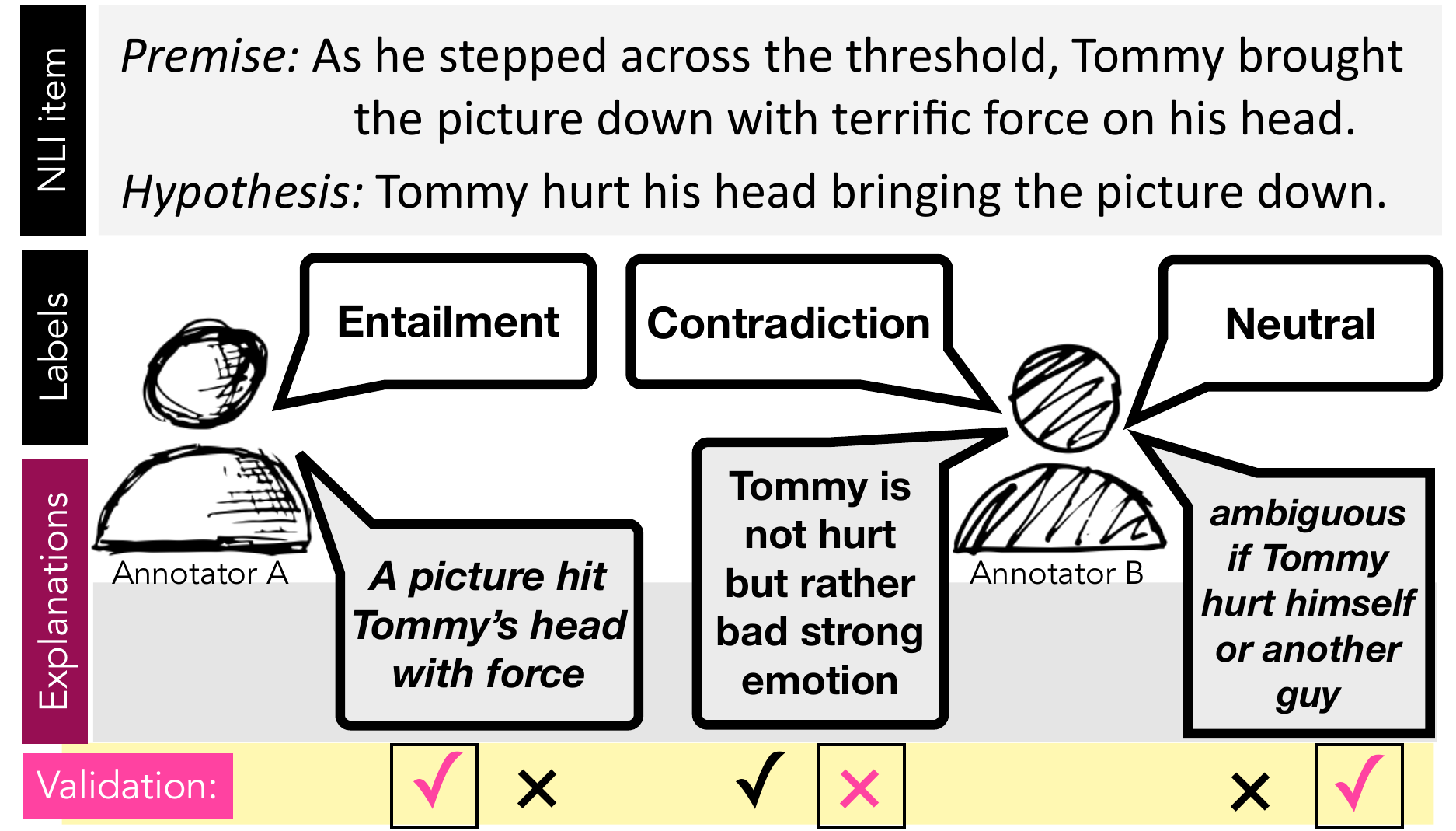}
\caption{Variation or Error? 
We present a procedure and multi-label dataset, \name, to tease apart annotation error from plausible human label variation. 
We leverage \textit{ecologically valid explanations} and \textit{validation} as two key mechanisms (boxed: self-validations; label ``Contradiction'' is an \emph{error}); see \S\ref{sec:dataset}-\S\ref{sec:validating-explanations-label} for details.
}
\label{fig:illustration}
\end{figure}

At the same time, there is increasing evidence that for many items in many tasks, more than a single label is valid. For some items, systematic variation 
exists for valid reasons, such as plausible disagreement or multiple interpretations. 
In other words, the world is not just black and white. Human label variation  (HLV, as termed by \citealt{plank-2022-problem}) has been shown on a wide range of NLP tasks~\cite{de-marneffe-etal-2012-happen,plank2014linguistically,aroyo2015truth}, including in natural language inference (NLI; \citealt{pavlick-kwiatkowski:TACL19,zhang-de-marneffe-2021-identifying}).
NLI involves determining whether a hypothesis is true (Entailment), false (Contradiction), or neither (Neutral), assuming the truth of a given premise; see Figure~\ref{fig:illustration} for an example with plausible labels.

Although high-quality, consistently labeled data may initially appear to conflict with the goal of accommodating HLV, it is important to emphasize that we do not perceive these as contradictory goals. While HLV exists, so do errors. We assert that annotators are inevitably prone to make errors, such as misunderstanding instructions or accidentally selecting a wrong label. Optimizing data quality is essential through providing clear instructions and effective training, and identifying annotation errors yields better datasets~\cite{larson-etal-2019-outlier}. However, still little is known about what constitutes an error versus plausible variation. We lack both a theory and operationalizable procedures to tease apart error from plausible HLV consistently and soundly.  
Some datasets with errors (and their corrections) exist, and there has been work on automatic error detection (AED). However, both have their limitations (\S\ref{sec:related_work}).
A crucial gap remains: a lack of examination in real-world scenarios where the signal is nuanced, not merely black-and-white. 

To address this gap, this paper contributes:
    (i)~\name, a novel multi-annotator English NLI dataset with both 
    plausible variation and detected errors. To the best of our knowledge, no such dataset exists yet.
    (ii)~A new methodology to detect errors:
    we collect multiple annotations, where each label comes with an ecologically valid explanation inspired by \citet{jiang-etal-2023-ecologically}, and propose to pair them with validity judgments to identify errors. 
    (iii)~Finally, we benchmark existing AED methods and GPTs in a challenging setup, where the task is to tease apart error from plausible human label variation.
    Our findings indicate that existing AED methods underperform humans and GPTs substantially on our self-validated \name{} NLI dataset.
    We release our data and code to facilitate uptake.\footnote{\url{https://github.com/mainlp/VariErr-NLI}}


\section{Related Work}\label{sec:related_work}

Labeled data is the fuel of machine learning, as it drives both learning and evaluation. 
Following a data-centric view, we focus on improving data quality over data quantity~\cite{motamedi2021datacentric,swayamdipta-etal-2020-dataset,zhang-etal-2021-learning-different,gordon2022jury}. 
We aim to bring together work on data quality from two ends: annotation error vs.\ human label variation.

\paragraph{Annotation Errors and AED} 
Several recent work found errors in widely used benchmarks, such as CoNLL 2003 for Named Entity Recognition~\cite{wang_crossweigh_2019, reiss_identifying_2020,rucker-akbik-2023-cleanconll}, TACRED for relation extraction~\cite{altTACREDRevisitedThorough2020}, WSJ for syntax~\cite{manning2011part,dickinsonDetectingErrorsPartofSpeech2003}, and ImageNet for object classification~\cite{beyer2020we,northcutt2021pervasive,vasudevan2022does}. 

AED has a long-standing tradition in NLP. Proposed methods range from early work that relies on variation-based methods positing that instances with similar surface forms tend to have the same label~\citep{dickinsonDetectingErrorsPartofSpeech2003,plank2014linguistically} to more recent model-based approaches that either exploit 
predictions~\cite{amiriSpottingSpuriousData2018,arazoUnsupervisedLabelNoise2019} or information derived from training dynamics~\cite{swayamdipta-etal-2020-dataset};
see~\citet{klie-etal-2023-annotation} for a 
survey on AED.

Flaggers and scorers for AED have been proposed~\cite{klie-etal-2023-annotation}. Flaggers detect errors by providing a hard decision on whether an instance is erroneous. Scorers, on the other hand, assign a score to each instance reflecting the likelihood of being an error, and the top-$n$ scoring instances are then corrected. Here, we focus on scoring methods to rank instances. 
Most of the AED work mentioned has limitations as they either rely on post-hoc mining of errors (and might therefore miss out on errors) in semi-automatic ways~\cite[e.g.,][]{reiss_identifying_2020}, or they inject synthetic noise which has been shown to result in datasets where errors are easy to spot~\cite{larson-etal-2019-outlier}. Instead of using synthetic noise, we focus on realistic setups and re-annotate data in ecologically valid ways. 

\paragraph{Human Label Variation (HLV)}  Recent studies have drawn attention to HLV in NLP~\cite[i.a.,][]{uma_learning_2021,plank-2022-problem}. HLV has been described as annotator disagreement, which is not just noise but also signal since a
sign of vagueness or ambiguity can benefit models \citep{aroyo2013crowd}.
These include judgments that are not always categorical~\cite{de-marneffe-etal-2012-happen}, inherent disagreement~\cite{pavlick-kwiatkowski:TACL19,davani2022dealing},
or justified and informative disagreement~\cite{sommerauer-etal-2020-describe}.
For subjective NLP tasks, which by essence encourage annotator subjectivity (and hence variation), there is also a line of work referred to as perspectivism~\cite{Cabitza_Campagner_Basile_2023},
connected to the descriptive data annotation framework proposed by~\citet{rottger-etal-2022-two}.

\paragraph{HLV in NLI} This paper focuses on NLI, known to contain HLV~\citep{pavlick-kwiatkowski:TACL19,nie-etal-2020-learn,jiang-marneffe-2022-investigating,jiang-etal-2023-ecologically}. 
\citet{pavlick-kwiatkowski:TACL19} re-annotated nearly 500 NLI instances with 50 crowd-workers and showed that disagreements in NLI cannot be dismissed as annotation ``noise.'' 
ChaosNLI \citep{nie-etal-2020-learn} pioneers large-scale NLI annotation by collecting 100 annotations per instance for 3K items from SNLI \cite{bowman-etal-2015-large}, $\alpha$NLI \citep{Bhagavatula2020Abductive}, and MNLI \citep{williams-etal-2018-broad} but 
for which the original annotations did not yield high agreement.
They show that, for most of the items, HLV persists with more annotations. Further, their experiments show a large room for model improvement and a positive correlation between human agreement and label accuracy.

In another line of work, \citet{jiang-marneffe-2022-investigating} identified \textit{reasons} for observing variation in NLI, deriving a taxonomy based on linguistic properties of the items. Following up on that work, \citet{jiang-etal-2023-ecologically} proposed \textsc{LiveNLI}, to gain insights into the origins of label variation. They re-annotated 122 NLI instances from ChaosNLI with \textit{ecologically valid explanations}: annotators are instructed to not only provide NLI labels but also explanations for their label choices. This addresses a limitation of prior work that uses post-hoc explanations, which may not reflect the true reasons of the original annotators, thereby questioning the validity of the prior method. 
They show that ecologically valid explanations have an additional benefit:  signaling \textit{within-label
variation}, i.e., annotators give the same label but for different reasons. While we do not focus on the latter here, we take inspiration from \citet{jiang-etal-2023-ecologically} to collect ecologically valid explanations (cf. \S\ref{subsec:round1}).  

To the best of our knowledge, there remains a gap for studies on \textit{both} annotation errors and human label variation in a concentrated effort. It is thus an open challenge to define error in an ecologically valid way, and it is unknown to what extent existing AED methods help detect such errors and whether new methods are needed. To find answers to these challenging open questions, we believe it is important to move both directions forward.


\begin{table*}[t]
\centering
\begin{subtable}[h!]{1.0\textwidth}
\resizebox{1.0\textwidth}{!}{
\begin{tabular}{ccl|cccc}
\multicolumn{7}{l}{\begin{tabular}[c]{@{}l@{}}
\textit{Premise}: 
\texttt{They made little effort, despite the Jesuit presence in Asia, to convert local inhabitants} \\
\texttt{to Christianity or to expand their territory into the interior.} \\
\textit{Hypothesis}: 
\texttt{The Jesuit presence in Asia helped to convert local residents to Christianity, } 
\\[0pt]
\texttt{allowing them to expand their territory.} \\
\textit{Label-explanation pairs}:
\texttt{Before:\{E:1,C:4\}} \;
\texttt{Self-validated:\{C:3\}} \;
\texttt{Peer-validated:\{C:4\}}  
\\
\textit{Label:} \texttt{[C]} \;
\texttt{Error:[\codebox{E}]} \;
\end{tabular}} \\
\hline
\multicolumn{3}{c|}{\textit{Round 1: NLI Label \& Explanation}}
& \multicolumn{4}{c}{\textit{Round 2: Validity}}  \\
\multirow{1}{*}{L}  
& \multirow{1}{*}{A} 
& \multicolumn{1}{c|}{Explanation}
& 1 & 2 & 3 & 4 \\
\hline
\multirow{1}{*}{\codebox{E}} & 1 & 
\texttt{Both premise and hypothesis suggest that the speaker does not understand.}
& \noself & \no & \no & \no
\\
\hline
\multirow{4}{*}{C} & 1
& 
\begin{tabular}[c]{@{}l@{}}\texttt{The Jesuit didn't make much effort to convert local residents to Christianity} 
\\[-2pt]
\texttt{or to expand their territory.}\end{tabular}
& \yesself & \yes & \yes & \yes
\\[-2pt]
& 2 & 
\texttt{They did not try to expand their territory.}
& \yes & \idkself & \yes & \yes
\\[-2pt]
& 3  &
\begin{tabular}[c]{@{}l@{}}\texttt{The Jesuit did not make effort to convert local residents to Christianity} 
\\[-2pt]
\texttt{or to expand their territory. }\end{tabular}
& \yes & \yes & \yesself & \yes
\\[-2pt]
& 4
& 
\begin{tabular}[c]{@{}l@{}}\texttt{They made little effort to convert the locals or to expand their territory. } 
\\[-2pt]
\texttt{So they did not help.}
\end{tabular}
& \yes & \yes & \yes & \yesself 
\\[-2pt]
\hline
\end{tabular}
}
\vspace{-4pt}
\caption{
\textit{id:} 28306c}
\label{subtab:sample_annotation_varierr_22}
\end{subtable}
\hfill
\begin{subtable}[h]{1.0\textwidth}
\resizebox{1.0\textwidth}{!}{
\begin{tabular}{ccl|cccc}
\multicolumn{7}{l}{\begin{tabular}[c]{@{}l@{}}
\textit{Premise}: 
\texttt{Because marginal costs are very low, a newspaper price for preprints might be as low as} \\
\texttt{5 or 6 cents per piece.} \\
\textit{Hypothesis}: 
\texttt{Newspaper preprints can cost as much as \$5.} \\
\textit{Label-explanation pairs}:
\texttt{Before:\{E:1,N:2,C:1\}} \;
\texttt{Self-validated:\{N:2\}} \;
\texttt{Peer-validated:\{N:2,C:1\}} 
\\
\textit{Label:} \texttt{[N]} \;
\texttt{Errors:[\codebox{E, C}]}
\end{tabular}} \\
\hline
\multicolumn{3}{c|}{\textit{Round 1: NLI Label \& Explanation}}
& \multicolumn{4}{c}{\textit{Round 2: Validity}}  \\
\multirow{1}{*}{L}  
& \multirow{1}{*}{A} 
& \multicolumn{1}{c|}{Explanation}
& 1 & 2 & 3 & 4 \\
\hline
\multirow{1}{*}{\codebox{E}} & 4 & 
\texttt{5 dollars for a piece of newspaper.}
& \no & \no & \no & \noself
\\
\hline
\multirow{2}{*}{N} & 1
& 
\begin{tabular}[c]{@{}l@{}}\texttt{The context only mentions how low the} 
\texttt{price may be, not how high it may be.}\end{tabular}
& \yesself & \yes & \yes & \yes
\\[-2pt]
& 3 & 
\begin{tabular}[c]{@{}l@{}}\texttt{The maximum cost of newspaper preprints}
\texttt{is not given in the context.}
\hspace{2cm}
\end{tabular}
& \yes & \yes & \yesself & \yes
\\
\hline
\codebox{C} & 2  
& 
\texttt{The context says 5 or 6 cents, not \$5.}
& \no & \noself & \yes & \yes
\\
\hline
\end{tabular}
}
\vspace{-4pt}
\caption{
\textit{id:} 72870c}
\label{subtab:sample_annotation_varierr_229}
\end{subtable}
\vspace{-4pt}
\caption{Sample \name{} NLI annotations. L:~Label, A:~Annotator; E:~Entailment, N:~Neutral, C:~Contradiction; 
\yes:~`yes (makes sense)'; 
\no:~`no';
\idk:~`IDK (I don't know)';
\fbox{\textcolor{magenta}{magenta}}: self-judgments, 
black: peer-judgments, 
\codebox{Err}: label error.
Curly brackets in \textit{label-explanation pairs} denote label counters, e.g., in \ref{subtab:sample_annotation_varierr_22}, \texttt{Before:\{E:1,C:4\}} means that there are one entailment and four contradiction label-explanation pairs before validation.
}
\label{tab:sample_annotation_varierr}
\end{table*}

\section{\name{}: Annotation Procedure}\label{sec:dataset}

To tease apart human label variation from error, we create \name{} (Variation versus Error), a NLI dataset with two rounds of annotations by four annotators:\footnote{Annotators are Master's students in Computational Linguistics and the first author of this paper, all paid according to national standards.}
Round 1 for NLI labels and explanations 
(\S\ref{subsec:round1})
and Round 2 for validity judgments
(\S\ref{subsec:round2}). 
Table \ref{tab:sample_annotation_varierr} presents two \name{} examples with two-round annotations, as well as their deduced label variations and errors to be discussed in \S\ref{sec:validating-explanations-label}.

\subsection{Round 1: NLI Labels \& Explanations}\label{subsec:round1}

We collect annotations from four annotators on 500 NLI items randomly sampled from the MNLI subset of ChaosNLI.
Annotators were asked to provide not only one or more NLI labels (E: Entailment, N: Neutral, C: Contradiction) to each item but also a one-sentence \emph{explanation} for each label they chose, as the same label could be chosen for different reasons \cite{jiang-etal-2023-ecologically}. Annotators could use a fourth ``I don't know'' (IDK) label if none of the NLI labels seemed suitable. 
The Round 1 annotation sums up to 1,933 label-explanation pairs with the standard three NLI labels for the 500 items,
and 331 ``IDK'' annotations (released in the data) which are discarded in Round 2.

\subsection{Round 2: Validity Judgments}\label{subsec:round2}

\name's key contribution lies in proposing a second round of \emph{validity judgment}.
Validity judgment mirrors conventional annotation adjudication in that annotators judge each other's NLI labels and explanations. 
This information is delivered anonymously to annotators to reduce group dynamics. 
However, rather than agreeing on a single label or explanation altogether, annotators are free to make independent judgments on each label-explanation pair from Round 1, which enables inferring what is an error versus plausible variation (cf.\ \S\ref{subsec:validating-explanations}). 

So in Round 2, annotators become judges. For all 500 items, the 1,933 label-explanation pairs from Round 1 are distributed anonymously to the same four annotators.
For each NLI item, 
each judge sees all label-explanation pairs annotated in Round 1, including their own, which they may or may not remember.\footnote{Round 1 took place August-November 2023, and each annotator spent $\sim$26 hours to annotate 500 items.
Round 2 took place from November 2023 to January 2024, and each spent $\sim$15 hours.
Annotators worked independently in different weeks. The interval between the end of Round 1 and the start of Round 2 was one month or longer for all annotators. }  
For each label-explanation pair, the annotator judges whether the explanation makes sense for the NLI label, answering 
``yes'' (\yes), ``no'' (\no) or ``IDK'' (\idk, I don't know) as shown in the four right columns of Table \ref{tab:sample_annotation_varierr}.
Round 2 amounts to 7,732 validity judgments, including 158 ``IDK''s.

\section{\name{}: Detecting Errors}\label{sec:validating-explanations-label}

Multiple validity judgments on label-explanations enable distinguishing annotation errors from HLV. 

\subsection{Self versus Peer}\label{subsec:self-vs-peer}
One consequential feature of our two-round multi-annotator procedure is the post-hoc distinction between self- vs.\ peer-judgments.
\textit{Self-judgments} refer to Round~2 judgments on the judge's own Round~1 label-explanation annotations ( \yesself{ }, \mbox{\noself{ },} \idkself{} in Table~\ref{tab:sample_annotation_varierr}), whereas \textit{peer-judgments} refer to judgments from other annotators (\yes, \no, \idk). 
Since we have four annotators, each label-explanation pair receives one self-judgment and three peer-judgments. 
Note that the self vs.\ peer distinction only enters into effect after data collection. 

\subsection{Validating Labels}
\label{subsec:validating-explanations}

Let $\mathcal{A}=\{a_1,..,a_4\}$ be the set of annotators.

\paragraph{Self-validated Label-Explanation}
A label-explanation pair given by annotator $a_k$ on an item 
in Round 1 is \emph{self-validated} if $a_k$
marks the label-explanation pair as ``yes'' in Round 2.

\paragraph{Peer-validated Label-Explanation}
A label-explanation pair given by annotator $a_k$ 
in Round 1 is \emph{peer-validated} if the majority ($\geq$2) of the other three annotators $\mathcal{A}\backslash \{a_k\}$ marks the pair as ``yes'' in Round 2.

For example, the item in  Table~\ref{subtab:sample_annotation_varierr_22} received the Contradiction (C) label and accompanying explanations from all four annotators in Round 1. 
Among these four label-explanations, three are \textit{self-validated} (~\yesself{ }) in Round 2.
On the other hand, all four explanations for C are \textit{peer-validated} since the majority (all in this case) of the peers voted ``yes'' (\yes{}) for each label-explanation.

\subsection{What counts as an error?}\label{subsec:what-count-as-an-error}

In the conventional setup of annotation adjudication, multiple annotators discuss the rationales for their labels and converge to an agreed label. 
The annotations that are originally different from and subsequently changed to the agreeing label are considered annotation errors. 
Similarly, in \name{}, a label-explanation pair might be considered wrong in retrospect (i.e., in Round 2) by the annotator who wrote it after reading all label-explanation pairs given to that item by all annotators.

Thus, in this paper, we use Round 2 \textit{self-judgments}---approving or rejecting annotators' own Round 1 label-explanation pairs---as the criteria for annotation errors.\footnote{We opted for a strict error definition here. Peer validation could be used in the future but requires additional decisions.} 
We define a NLI label as an \textit{error} if all label-explanation pairs are not self-validated.
In other words, a label is viewed as correctly attributed to an item if any of its explanations is self-validated. 
In Table~\ref{subtab:sample_annotation_varierr_22}, the Contradiction (C) label has at least one self-validated \mbox{(~\yesself{ })} explanation (it even has three), and is thus not deemed an error. 
In contrast, Entailment (\codebox{E}) is an error in Table~\ref{subtab:sample_annotation_varierr_22} because none of its explanations is self-validated, similarly for \codebox{E} and \codebox{C} in Table~\ref{subtab:sample_annotation_varierr_229}.

\subsection{Data Statistics \& IAA}
Table~\ref{tab:inter-annotator-agreement} shows the frequencies of NLI labels across the four annotators on the 500 items and 1,933 explanations before and after validation. 
We include statistics on \emph{repeated} frequency counts (e.g., E counts twice if it is given as a label by two annotators for the same item) and \emph{aggregated} labels (repeated labels for a given item count once). Moreover, following \citet{jiang-etal-2023-ecologically}, we compute inter-annotator agreement (IAA) on NLI labels using Krippendorff’s $\alpha$ (for multi-annotator) with MASI-distance (for multi-label).

\begin{table}[t]
\centering
\resizebox{0.48\textwidth}{!}{
\begin{tabular}{c|lcccr|c}
\toprule
Validation & FreqType & \textbf{E} & N & C & $\sum$  & IAA \\
\midrule
\multirow{2}{*}{before validation}  & 
\textit{repeated} & 554 & 977 & 402 & 1,933
& \multirow{2}{*}{0.35} \\
& \textit{aggregated} & 263 & 403 & 212 & 878  &  \\
\midrule
\multirow{2}{*}{self-validated} 
& \textit{repeated} & 467 & 916 & 329 & 1,712
& \multirow{2}{*}{0.50} \\
& \textit{aggregated} & 210 & 380 & 159 & 749 &  \\
\midrule
\multirow{2}{*}{peer-validated} 
& \textit{repeated} & 446 & 859 & 296 & 1,601 
& \multirow{2}{*}{0.69} \\
& \textit{aggregated} & 177 & 335 & 130 & 642 &  \\
\bottomrule
\end{tabular}
}
\caption{Frequency counts and inter-annotator agreement (Krippendorff’s $\alpha$ with MASI-distance) on non-, self-, and peer-validated \name \ NLI labels.}
\label{tab:inter-annotator-agreement}
\end{table}

\begin{figure*}[t]
\begin{subfigure}[h!]{0.39\textwidth}
\centering
\includegraphics[width=\textwidth]{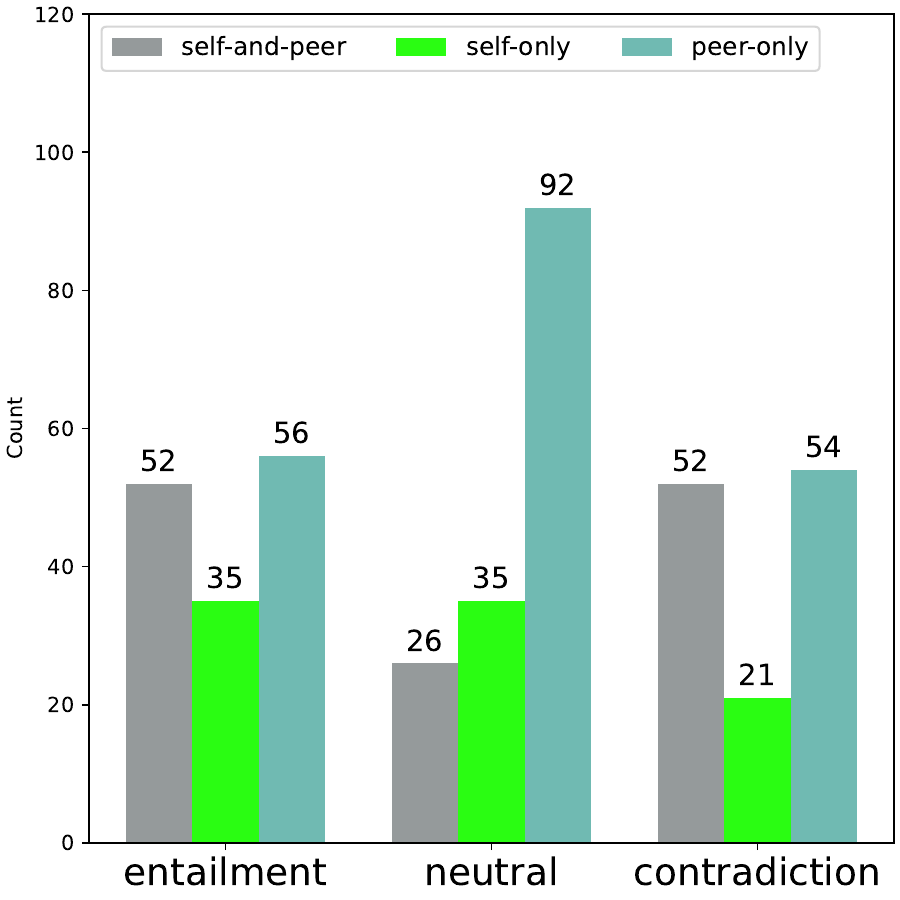}
\caption{Number of label-explanations rejected by self-and-peer, self-only, and peer-only validations.}
\label{subfig:barplot-label-sets-by-label}
\end{subfigure}
\hfill\hfill
\begin{subfigure}[h!]{0.59\textwidth}
\centering
\includegraphics[width=\textwidth]{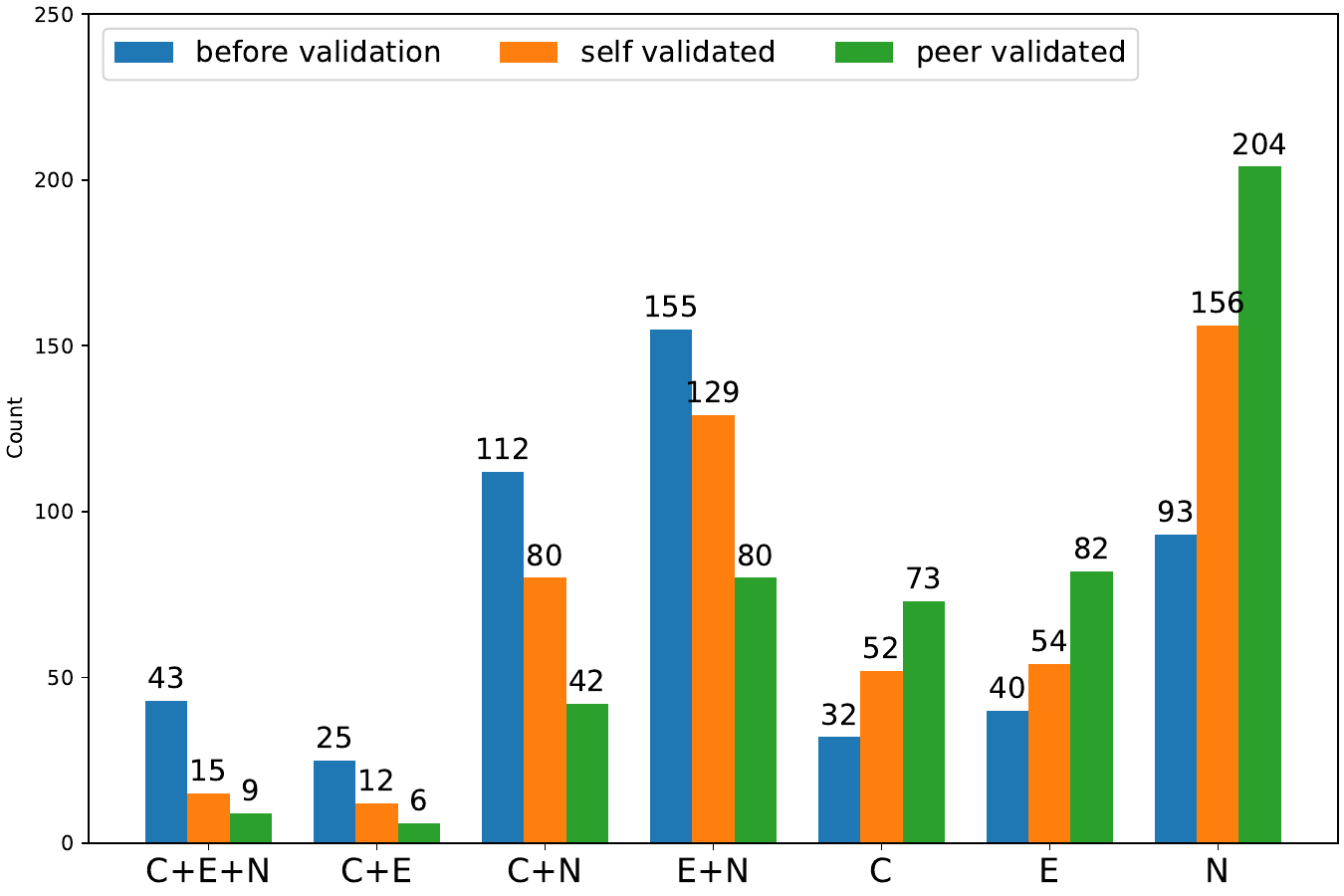}
\caption{NLI label sets on non-, self- and peer-validated items.}
\label{subfig:barplot-label-sets-per-item}
\end{subfigure}
\label{fig:barplot-stats}
\caption{Frequency statistics on \name.}
\end{figure*}

Since all \name{} items are sampled from ChaosNLI, which only includes MNLI items with two or all three of the NLI labels in the original annotations by design, we expect HLV and, thus, a medium-to-low IAA in our dataset.
Indeed, \name{} has an IAA of 0.35 (Krippendorff’s $\alpha$ with MASI-distance) before validation, which raises to 0.50 and 0.69 after self- and peer-validations, with the latter reaching substantial agreement  (see \ref{subsec:appx-pairwise-iaa} for pairwise IAA).
However, with the appreciation of HLV, as long as there are no errors in the data, we argue that perfect agreement is not reachable. As a matter of fact, though the repeated and aggregated frequencies of NLI labels decrease adequately after validation, 
HLV still exists in self- and peer-validated annotations, averaging 1.50 (749/500) and 
1.28 (642/500) labels/item.

We also observe in Table~\ref{tab:inter-annotator-agreement} that 88.57\% (1,712/1,933) of Round 1 explanations in \name{} were self-validated and 82.82\% (1,601) were peer-validated.
Figure~\ref{subfig:barplot-label-sets-by-label} presents the number of label-explanation pairs rejected by both self- and peer-validations, by self-validation only, and by peer-validation only. 
Most Entailment and Contradiction annotations rejected by self are also rejected by peers (dark green).
However, Neutral presents a challenging situation for self-validation where 60.13\% (92/153) of Ns are only invalidated by the joint force of peers but not by one annotator alone. 

Figure~\ref{subfig:barplot-label-sets-per-item} gives the frequencies of aggregated label combinations per item before validation and after self- and peer-validations 
(see~\ref{subsec:appx-barplot-label-sets-per-explanation} for label-explanation pair frequencies). 
Frequencies of multi-labeled items drop after self-validation and, more remarkably, after peer-validation. 
Inversely, the number of single-labeled items increases vastly, especially for Neutral.
We also observe from \name{} that 
a large portion of items, 37.6\% (188/500), are self-identified as errors and 
51.6\% (258) are rejected by peer-validation.

In sum, though HLV remains in \name{}, 
our validation process demonstrates that annotation errors are frequently concealed under label variations. 
We thus proceed with the challenging automatic error detection task in \S\ref{sec:aed}-\ref{sec:results} to separate annotation errors from valid HLVs.

\section{Automatic Error Detection (AED) on \name}\label{sec:aed}

We now describe our experiments to detect annotation errors using \name{} automatically. 
We evaluate the capabilities of AED methods, LLMs, and human heuristics (all henceforth \emph{scorers}) in capturing annotation errors.

\subsection{Task Definition and Evaluation}

Following~\citet{klie-etal-2023-annotation} and \citet{weber-plank-2023-activeaed}, we model AED as a ranking task.
In this setting, the goal of the \textit{scorer} is to provide a ranked list with the labels that are most likely errors at the top and the most likely correct ones at the bottom.
This ranked list can then be used to guide re-annotation efforts~\citep{altTACREDRevisitedThorough2020,northcutt2021pervasive} or remove the most likely errors from the training data~\citep{huangO2UNetSimpleNoisy2019}.
Scorers produce such a list by assigning an error score to each assigned label in the dataset.
They derive the ranked list by sorting it based on the assigned scores.

We evaluate scorers on \name{} using the following protocol.
A model receives the list of NLI items from \name{} where each item is paired with the \textit{aggregated} label(s) it received in Round 1.  
For the 500 items in \name{}, the model is given a list of 878 item-label pairs (cf.\ Table \ref{tab:inter-annotator-agreement}).
Based on that information, the model assigns an error score and ranks the labels by this score.
We evaluate how well the model performs by comparing this ranked list with the self-flagged errors (\S\ref{subsec:what-count-as-an-error}).
Following~\citet{klie-etal-2023-annotation}, we use standard ranking metrics for evaluation: average precision (AP), i.e., the area under the precision/recall curve computed over all assigned labels, and precision/recall for the top 100 ranked labels, P@100 and R@100.

\subsection{Models}
We evaluate five different AED models: 
two variants of Datamaps (DM, \citealt{swayamdipta-etal-2020-dataset}),
Metadata Archaeology (MA, \citealt{siddiqui-et-al-2023-metadata}),
and two GPTs.
We report the mean and standard deviation over three random seeds for DM and MA.

\paragraph{Datamaps (DM)}
We use training dynamics (i.e., the collection of training statistics over epochs $E$) for each label. These statistics are obtained by training a DistilRoBERTa-base model\footnote{\url{https://huggingface.co/distilroberta-base}}~\citep{Sanh2019DistilBERTAD} following~\citet{klie-etal-2023-annotation} in a multi-label setting~\citep{jiang-marneffe-2022-investigating} on all labels of \name{} obtained in Round 1.
We refer to the $j$'th label of the $i$'th example as $\text{label}_{i,j}$.
The training dynamics are modeled by the probability $p_{i,j,e}$ that DistilRoBERTa predicts for $\text{label}_{i,j}$ after the $e$'th epoch.
Based on these probabilities, the two DM models we use are defined as follows:
\begin{align}
     \textit{DM}_{\text{mean}} &= - \frac{1}{E} \sum_{e=1}^{E} p_{i,j,e} \\
     \textit{DM}_{\text{std}} &= \sqrt{\frac{1}{E} (\sum_{e=1}^{E} p_{i,j,e} + \textit{DM}_{\text{mean}})^2}
\end{align}

Note that a \textit{low} average probability for the label indicates a likely error.
Because our evaluation setup requires the most likely errors to be ranked first, we negate the average probabilities.

\paragraph{Metadata Archaeology (MA)}
MA models AED as a supervised task. It represents each instance (or label in our case) as the $E$-dimensional $-\log{p_{i,j,e}}$ vector, where $E$ is the number of epochs and $p_{i,j,e}$ is the probability the model assigns to the $j$'th label of the $i$'th NLI instance at epoch $e$.
Then, it assumes that some instances are labeled with the property of interest (in our case, whether it is an erroneous label).
It predicts whether an instance is an error by employing a k-nearest neighbors (kNN) classifier using the instance representations and error labels.
We use the average number of annotated errors for the kNN to obtain a score for each instance.
Following \citet{siddiqui-et-al-2023-metadata}, we use $k = 20$.
To obtain unbiased predictions, we require that the kNN training instances are distinct from those we want to obtain predictions for.
%
We use a 2-fold cross-validation setup where we split \name{} into two folds, use one half as ground truth, obtain the predictions for the other, and vice versa.

\paragraph{GPT}
We also compare two large language models (LLMs): GPT-3.5~\citep{brown-et-al-2020-language} and GPT-4~\citep{openai-2023-gpt4}.\footnote{GPT models have previously seen the premise-hypothesis pairs from MNLI (\citealt{balloccu-etal-2024-leak} and \url{https://hitz-zentroa.github.io/lm-contamination/}), but not the ample new annotations, i.e., NLI labels, explanations, and self/peer-validation judgments from multiple annotators.}
We emulate the Round 2 annotation setting in \S\ref{subsec:round2} as closely as possible by prompting each model to provide a score reflecting how much each Round 1 explanation makes sense for a given label.
We compute the score per label by averaging the GPT-assigned scores of all explanations for the label.
We prompt GPT as follows, giving it the premise (context) and hypothesis (statement) of a NLI item as well as all label-explanation pairs, asking it then to provide a probability for each reason:

\begin{lstlisting}
System: 
You are an expert linguistic annotator.

User:
We have collected annotations for a NLI instance together with reasons for the labels. Your task is to judge whether the reasons make sense for the label. Provide the probability (0.0 - 1.0) that the reason makes sense for the label. Give ONLY the reason and the probability, no other words or explanation. For example:

Reason: <The verbatim copy of the reason>
Probability: <the probability between 0.0 and 1.0 that the reason makes sense for the label, without any extra commentary whatsoever; just the probability!>.

Context: {CONTEXT}
Statement: {STATEMENT}

Reason for label {LABEL}: {REASON_1}
Reason for label {LABEL}: {REASON_2}
[...]
Reason for label {LABEL}: {REASON_n}

Reason {REASON_1}
Probability: 
\end{lstlisting}

We implement GPTs using sglang~\citep{zheng2023efficiently} and its default sampling parameters.
See Appendix \ref{sec:appx-gpt-prompt} for a complete prompt example.
Note that the GPTs have access to the explanations for the labels, whereas the other models described above only have access to the labels without explanations.

\subsection{Human Heuristics}
In addition to the above automatic means, we experiment with four human heuristics that use the human label distributions over NLI labels (E, N, C) from annotation efforts:
label counts from ChaosNLI (100 annotators) and \name{} (4 annotators). In addition, we compare to  \name{}'s total and average peer judgments over explanations.

\paragraph{Label Counts (LC): ChaosNLI \& \name}
We hypothesize that if multiple annotators choose the same label, there is a high likelihood that the label is a correct annotation. 
We implement two label count (LC) baselines: one using ChaosNLI  \citep{nie-etal-2020-learn} and one using \name.\footnote{We did not include a comparison with LiveNLI \citep{jiang-etal-2023-ecologically} because among its re-annotated 122 MNLI items, only 15 are shared with \name{}.}
Since \name{} is a subset of ChaosNLI items,
we use label counts from ChaosNLI (LC\textsubscript{\textsc{Chaos}}) as a human heuristic to score Round 1 labels on each item, i.e., how many of the 100 crowd-workers annotated $\text{label}_{i,j}$ on item $i$. 
For instance, the ChaosNLI human distribution is \{N:25, E:72, C:3\} for the example in Figure~\ref{fig:illustration}.
Similarly, we include LC\textsubscript{\name{}} that counts the number of annotators (4 in total) that assigns $\text{label}_{i,j}$ to item $i$ in \name's Round 1 NLI labels. 
We multiply both LC\textsubscript{\textsc{Chaos}} and LC\textsubscript{\name{}} by $-1$, proposing that if a label has a higher count, then it is less likely to be an error.

\paragraph{Peer-judgments (Peer) in \name}
\name's 2-round annotations enable more fine-grained human heuristics that engage judgments on label-explanation pairs. 
Since each $\text{label}_{i,j}$ can be assigned by multiple annotators with different explanations, we count the number of ``yes'' judgments on explanations from peers, i.e., excluding self-judgments since those are used for gold error labels.

We implement two peer heuristics: Peer\textsubscript{sum} and Peer\textsubscript{avg}.
Peer\textsubscript{sum} sums all ``yes'' judgments across multiple explanations on the same label,
and
Peer\textsubscript{avg} sums ``yes''  judgments within each explanation and then averages across explanations within the label.
Given that one label can maximally receive four explanations, it can receive up to 12 peer-judgments, 3 per explanation.
For example, C in Table \ref{subtab:sample_annotation_varierr_22} receives 11 peer-judged ``yes'' in total (Peer\textsubscript{sum}$=3+2+3+3=11$), and the average over four explanations is Peer\textsubscript{avg}$=11/4=2.75$.
Peer\textsubscript{avg} differentiates more from Peer\textsubscript{sum} when there are multiple explanations, but each receives sparse ``yes'' judgments.
For example, N in Table \ref{subtab:appx_annotation_varierr_704} (Appendix \ref{sec:appx-more-examples}) receives 3 ``yes'' judgments but across two explanations, resulting in Peer\textsubscript{sum}$=2+1=3$ and Peer\textsubscript{avg}$=3/2=1.5$.
Similarly to the label counts above, we multiply both Peer\textsubscript{sum} and Peer\textsubscript{avg} by $-1$, hypothesizing that fewer ``yes'' judgments indicate a higher likelihood to be an annotation error.  

\paragraph{Combining Label Counts and Models}
Ranking labels by the number of annotations they received in Round 1 is a very strong baseline; see $\text{LC}_\text{\name}$ in Table~\ref{tab:results}.
Inspired by~\citet{frassetto-et-al-2019}, we investigate an approach that re-ranks the predictions of $\text{LC}_\text{\name}$ by breaking ties with the scores produced by another model (e.g., DM, MA or GPTs).
Note that $\text{LC}_\text{\name}$ produces many ties because its score is always one of $\{-1, -2, -3, -4\}$.

\section{Results for AED on \name{}}\label{sec:results}
Table~\ref{tab:results} presents human and model performances on \name{} AED using the ranking setup in \S\ref{sec:aed}.

\begin{table}[htbp]
\centering
\resizebox{0.49\textwidth}{!}{
\begin{tabular}{llll|l}
\toprule
Scorer & AP   &P@100&R@100   &AP (rerank)\\
\midrule
\multicolumn{5}{c}{\textit{Baselines}}  \\
Random & $14.7$&$14.7$&$11.4$ &-\\
\midrule
\multicolumn{5}{c}{\textit{Models}} \\
MA& $17.7\pm1.5$&$18.3\pm4.2$&$14.2\pm3.2$ &$44.2\pm3.0$\\
$\text{DM}_{\text{mean}}$& $22.8\pm0.4$&$23.7\pm2.1$&$18.3\pm1.6$ &$\textbf{50.4}\pm0.7$\\
$\text{DM}_{\text{std}}$& $22.3\pm1.9$&$22.7\pm1.2$&$17.6\pm0.9$ &$50.0\pm1.5$\\
GPT-3.5 & $17.6$&$21.0$& $16.3$ &$37.6$ \\
\textbf{GPT-4} & $\textbf{31.3}$&$\textbf{46.0}$& $\textbf{35.9}$ &$47.4$\\
\midrule
\multicolumn{5}{c}{\textit{Human}} \\
LC\textsubscript{\textsc{Chaos}} & $32.5$&$35.0$&$27.3$   &$\textbf{49.8}$\\
LC\textsubscript{\name} & 	$40.8$& $42.0$& $32.6$ &$40.8$\\ 
Peer\textsubscript{avg} & $42.2$&$46.0$&$35.9$ &$47.8$ \\
\textbf{Peer\textsubscript{sum}} & $\textbf{46.5}$&$\textbf{47.0}$&$\textbf{36.7}$ &$47.8$ \\
\bottomrule
\end{tabular}%
}
\caption{Results for AED on \name. AP: average precision;
rerank denotes using the method to break ties in LC\textsubscript{\name}. For MA and DM, we report mean and standard deviation over three random seeds. Note that GPTs have access to explanations.
}
\label{tab:results}%
\end{table}%

\subsection{Human Performance}
The best human heuristic is from peers (Peer\textsubscript{sum}), reaching a performance of $46.5\%$ AP, $47\%$ precision@100, and $36.7\%$ recall@100. 
These numbers support our hypothesis that human validation 
can be used as a strong means to detect annotation errors in a task with relatively high HLV because self- and peer-rejected label-explanation pairs overlap considerably (cf.\ Figure~\ref{subfig:barplot-label-sets-by-label}).
Interestingly, both peer-derived heuristics from \name{} perform better than  $\text{LC}_\textsc{Chaos}$ (3 linguists versus 100 crowd-workers), which suggests that having few highly-trained expert annotators is sufficient for reliable error detection, outperforming a larger number of crowd-workers.
$\text{LC}_\text{\name}$ outperforming $\text{LC}_\textsc{Chaos}$ on all metrics strengthens this finding. 
Next we compare humans to automatic means.

\subsection{Model Performance}

\begin{figure}[t]
\centering
\includegraphics[width=\columnwidth]{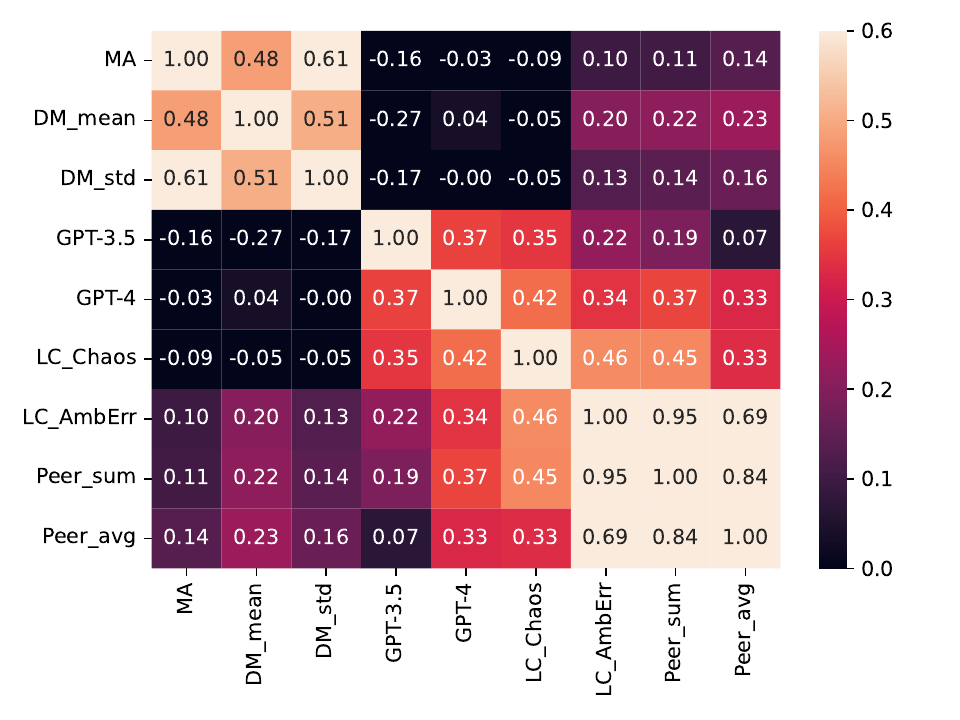}
\caption{Correlations among scorer predictions.}
\label{fig:correlation-heatmap}
\end{figure}

Among the models, GPT-4 outperforms all other methods by a large margin with a $8.5 / 22.3 / 17.6$ percentage points (pp.) improvement in terms of AP / P@100 / R@100 over the second best model $\text{DM}_\text{mean}$.
GPT-4 even outperforms LC\textsubscript{\textsc{Chaos}} in P@100 and R@100 and is close to the best peer heuristic for these two metrics.

One might postulate that ChaosNLI could have been part of GPT-4's training mixture, and GPT-4 performed well by reproducing its probabilities.
To check whether this is the case, we compute Pearson's $r$ between the predictions of all scorers (Figure~\ref{fig:correlation-heatmap}).
While GPT-4 has a slightly higher correlation ($0.42$) with $\text{LC}_\textsc{Chaos}$ than with all other methods, it is still much lower than some correlations between other models, e.g., $0.61$ between $\text{DM}_\text{std}$ and MA.
Thus, we conclude that GPT-4 does not solely rely on information from ChaosNLI but achieves its strong performance via some other mechanism.
Another possible explanation is that it is the only model next to GPT-3.5 that has access to explanations. In the future, we would like to investigate the use of explanations further.
Moreover, Figure~\ref{fig:correlation-heatmap} allows for a more general interesting observation. 
There seems to be a clear cluster structure in which the training-dynamics-based models (DM and MA) correlate highly with each other and GPT-4 clusters with human scorers.
Notably, correlations across these two clusters are small to non-existent or sometimes even negative.

\subsection{Influence of Human Label Variation}

\begin{figure}
    \centering
    \includegraphics[width=0.95\columnwidth]{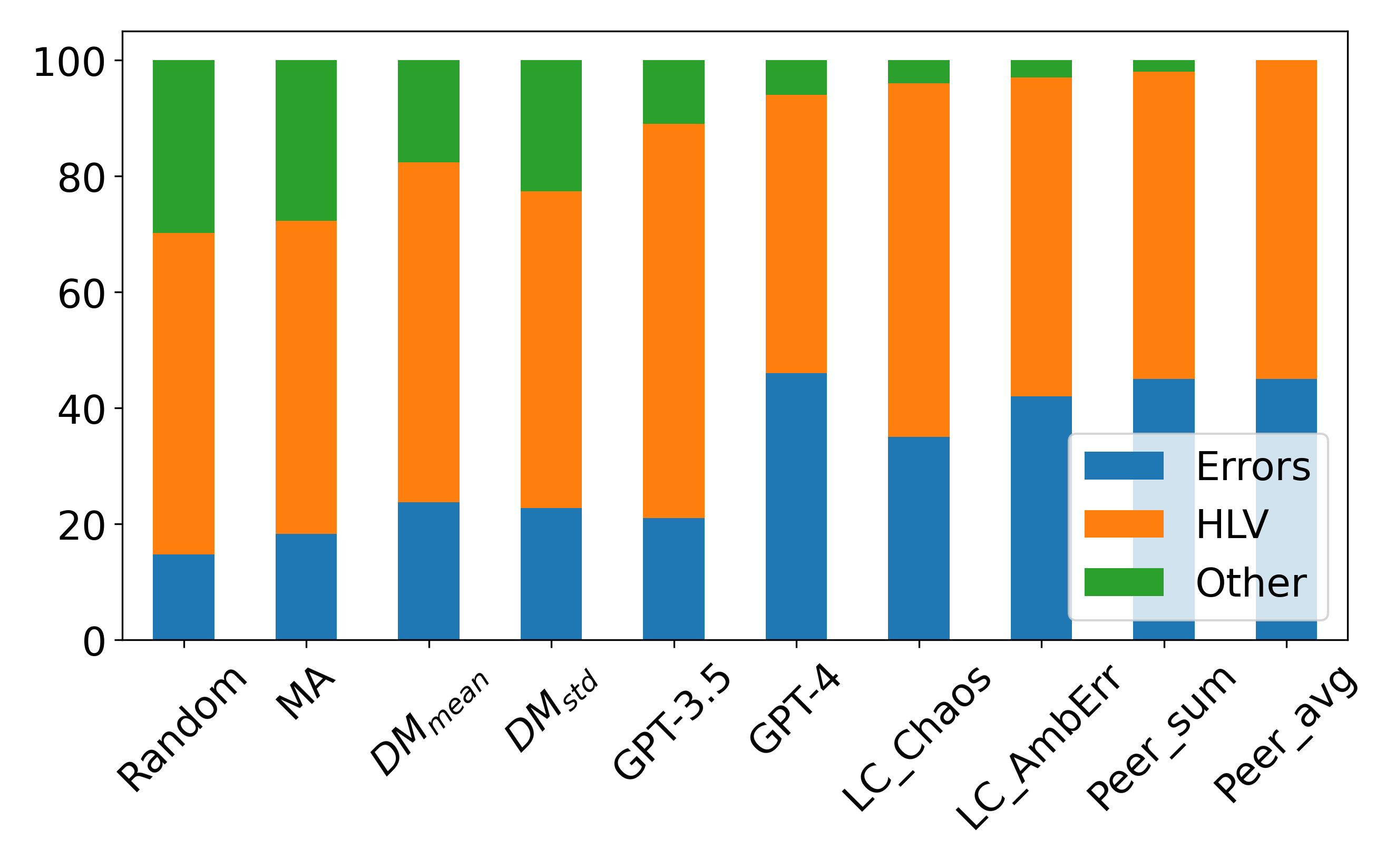}
    \caption{Average distribution of erroneous, HLV, and other labels over the top 100 instances per method.}
    \label{fig:instance-dist}
\end{figure}

In which situations do AED methods make mistakes, such as detecting false positive errors? 
This is an open question.
We hypothesize that many top-ranking labels would either be errors or come from instances displaying HLV, i.e., instances with multiple labels after self-validation. The rest should be instances with just one plausible label.
To test this hypothesis, we compute the proportion of \textit{erroneous labels} vs.\ \textit{valid labels from HLV instances} vs.\ \textit{other} (with a single plausible label and thus exhibiting neither errors nor HLV labels) for the top 100 ranking labels for each method. 

The results for the  GPTs and human heuristics in Figure~\ref{fig:instance-dist}
confirm our hypothesis: they place very few (0-11) labels that are neither errors nor HLV in the top 100.
On the other hand, the training-dynamics-based methods MA and DM assign between 17.6 and 29.8 of these items to the top 100.
This suggests that increasing the separation between errors and HLV is only one part of improving training dynamics methods for AED. Another could be finding the characteristics of the top-ranking items that are neither errors nor HLV.

\subsection{Reranking models using label counts}
Column \textit{AP (rerank)} in Table~\ref{tab:results} presents our re-ranking results.
We observe that re-ranking improves over vanilla $\text{LC}_\text{\name}$ for all methods but GPT-3.5.
Interestingly, the best performing methods---also compared to the non-re-ranking approaches---are $\text{DM}_\text{mean}$ and $\text{DM}_\text{std}$.
They even perform better than $\text{Peer}_\text{sum}$, the best human approach.
This suggests that combining statistics from multiple annotators with AED methods based on training dynamics is a promising future direction.

\section{Conclusion}\label{sec:conclusion}
Errors exist in datasets, but so does plausible human label variation. This paper defines a general procedure to separate the two by leveraging ecologically valid explanations (where annotators provide their reasons for a label) 
and pairing these with annotators' validations (to allow corrections). 
We provide a new \name{} dataset for the task of NLI re-annotated from scratch. 
Our empirical investigation on \name{} for NLI finds that traditional annotation error detection methods fare poorly and underperform humans and LLMs.

While this paper only applies our 2-round annotation procedure, \name{}, to NLI data, our methodology is general, and we hope it inspires uptake.
Future work includes adapting these approaches to other NLP tasks, probing differences between self- and peer-judgments, mapping such strategies to (large) language models, and linking \name{} to experiments with LLMs' 
explainability, self-correction, or multi-agent systems.

\section*{Limitations}
We believe that our two-round annotation setup would work for eliciting ecologically valid error annotations in tasks or languages other than English NLI.
However, we cannot be sure without trying it, which we did not do in this project.
Further, we did not use all types of information that \name{} contains for the training-dynamics-based AED methods.
An interesting question would be whether exploiting the soft label distribution with methods from learning from disagreement~\citep{uma_learning_2021} would improve AED results.
Another potentially useful source of information is the explanations given by the annotators.
Using this information for computing the training dynamics or directly modeling whether an explanation makes sense for a label in a supervised setting could potentially improve AED performance.

\section*{Acknowledgements}
We thank Huangyan Shan, Shijia Zhou, and Zihang Sun for their contributions and invaluable feedback on \name{}. Thanks also to Verena Blaschke for giving feedback on earlier drafts of this paper, as well as to the reviewers for their feedback.  Marie-Catherine de Marneffe is a Research Associate of the Fonds
de la Recherche Scientifique – FNRS.
This work is funded by ERC Consolidator Grant DIALECT 101043235 and supported by project KLIMA-MEMES funded by the Bavarian Research Institute for Digital Transformation (bidt), an institute of the Bavarian Academy of Sciences and Humanities. The authors are responsible for the content of this publication.

\bibliography{main,mypaper}

\newpage
\appendix

\section{Data Statistics}\label{sec:appx-data-statistics}

\subsection{Pair-wise inter-annotator agreements (Cohen's kappa) with MASI-distance for non-validated, self-validated, and peer-validated versions}\label{subsec:appx-pairwise-iaa}
\begin{table}[ht]
\centering
\resizebox{0.49\textwidth}{!}{
\begin{tabular}{l|cccccc}
\toprule
\textit{versions \textbackslash\ annotators} & 1-vs-2 & 1-vs-3 & 1-vs-4 & 2-vs-3 & 2-vs-4 & 3-vs-4 \\
\midrule
before validation & 0.40 & 0.42 & 0.37 & 0.36 & 0.31 & 0.34 \\
self-validated & 0.60 & 0.53 & 0.61 & 0.44 & 0.47 & 0.47 \\
peer-validated & 0.66 & 0.72 & 0.67 & 0.64 & 0.68 & 0.68 \\
\bottomrule
\end{tabular}
}
\caption{Pair-wise inter-annotator agreements (Cohen's kappa) with MASI-distance for non-validated, self-validated, and peer-validated versions.}
\label{tab:pairwise-iaa}
\end{table}

\subsection{Frequency of NLI label on non-validation, self-validated, and peer-validated explanation-label pairs}\label{subsec:appx-barplot-label-sets-per-explanation}
\begin{figure}[ht]
\centering
\includegraphics[width=0.49\textwidth]{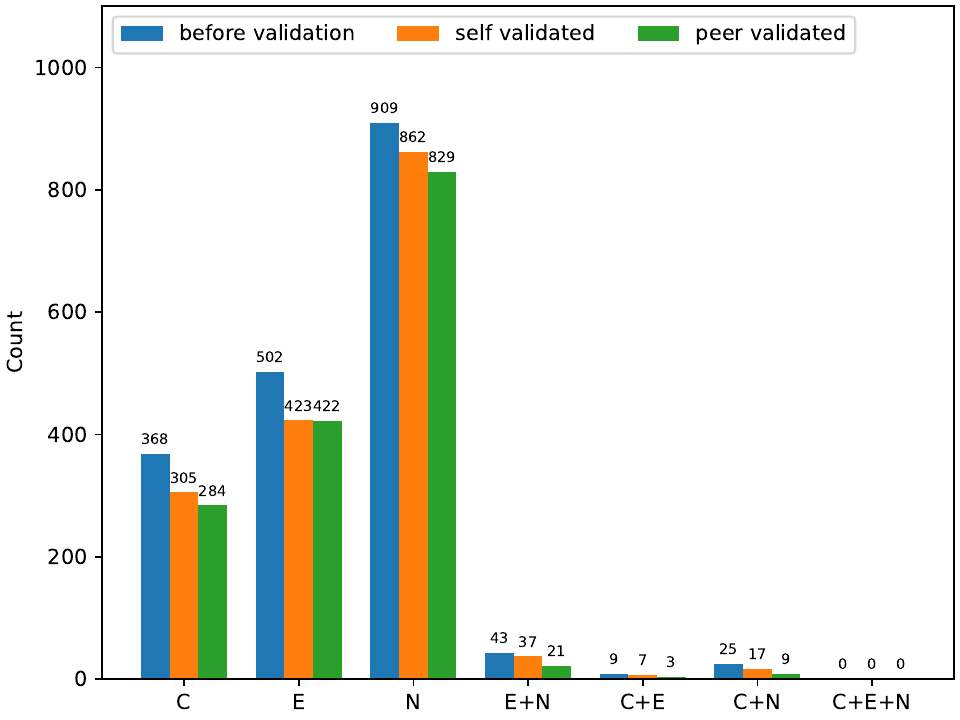}
\caption{Frequency of NLI label sets on non-, self- and peer-validated label-explanation pairs.}
\label{fig:appx-barplot-label-sets-per-explanation}
\end{figure}

\newpage
\section{GPT Prompt}\label{sec:appx-gpt-prompt}
\begin{figure}[ht]
\begin{minipage}[t]{0.49\textwidth} 
\begin{tcolorbox}[colback=gray!5,colframe=green!40!black,title=
\textit{id: }72870c]
\small
\textbf{System}: 
You are an expert linguistic annotator.

\textbf{User}: 
We have collected annotations for a NLI instance together with reasons for the labels. \\
Your task is to judge whether the reasons make sense for the label.  \\
Provide the probability (0.0 - 1.0) that the reason makes sense for the label.  \\
Give ONLY the reason and the probability, no other words or explanation.  \\
For example:  \\
\textit{Reason}: <The verbatim copy of the reason> \\
\textit{Probability}: <the probability between 0.0 and 1.0 that the reason makes sense for the label, without any extra commentary whatsoever; just the probability!>. \\
\textit{Context}: Because marginal costs are very low, a newspaper price for preprints might be as low as 5 or 6 cents per piece. \\
\textit{Statement}: Newspaper preprints can cost as much as \$5. \\
\textit{Reason for label entailment}: 5 dollars for a piece of newspaper \\
\textit{Reason for label neutral}: The context only mentions how low the price may be, not how high it may be. \\
\textit{Reason for label neutral}: The maximum cost of newspaper preprints is not given in the context. \\
\textit{Reason for label contradiction}: The context says 5 or 6 cents, not \$5. \\
\textbf{User}:
\textit{Reason}: 5 dollars for a piece of newspaper \\
\textit{Probability}: \\
\textbf{Assistant}:
0.0 \\
\textbf{User}: 
\textit{Reason}: The context only mentions how low the price may be, not how high it may be. \\
\textit{Probability}: \\
\textbf{Assistant}:
0.9 \\
\textbf{User}: 
\textit{Reason}: The maximum cost of newspaper preprints is not given in the context. \\
\textit{Probability}: \\
\textbf{Assistant}:
0.8 \\
\textbf{User}: 
\textit{Reason}: The context says 5 or 6 cents, not \$5. \\
\textit{Probability}: \\
\textbf{Assistant}: 
0.9
\end{tcolorbox} 
\end{minipage}  
\caption{GPT Prompt for predicting likelihood probability of label-explanation pairs.}
\label{fig:gpt-prompt-example}
\end{figure}

\section{More \name{} Examples}\label{sec:appx-more-examples}
\begin{table*}[t]
\centering
\begin{subtable}[h!]{1.0\textwidth}
\resizebox{\textwidth}{!}{
\begin{tabular}{ccl|cccc}
\multicolumn{7}{l}{\begin{tabular}[c]{@{}l@{}}
\textit{Premise}: 
\texttt{Students of human misery can savor its underlying sadness and futility.} \\
\textit{Hypothesis}: 
\texttt{Students of human misery will be delighted to see how sad it truly is.} \\
\textit{Label-explanation pairs}:
\texttt{before validation: \{E:1,N:2,C:1\}} \;
\texttt{Self-validated : \{E:1,N:1\}} \;
\texttt{Peer-validated: \{N:1\}} \\
\textit{Labels:} [E, N]
\quad \textit{Error:} [\codebox{C}]
\end{tabular}} \\
\hline
\multicolumn{3}{c|}{\textit{Round 1: NLI Label \& Explanation}}
& \multicolumn{4}{c}{\textit{Round 2: Validity}}  \\
\multirow{1}{*}{L}  
& \multirow{1}{*}{A} 
& \multicolumn{1}{c|}{Explanation}
& 1 & 2 & 3 & 4 \\
\hline
E & 2 & 
\texttt{``can savor'' implies ``will be delighted''.}
& \yes & \yesself & \no & \no
\\
\hline
 \multirow{2}{*}{N}  & 1 & 
\texttt{It is not clear from the context if the students will be delighted.}
& \noself & \no & \yes & \yes
\\
 & 3  & 
\begin{tabular}[c]{@{}l@{}}
\texttt{Students of human misery can ``savored'' that sadness, so maybe they are delighted} \\
\texttt{to see that, maybe they are tortured by the disasters.}
\end{tabular}
& \no & \no & \yesself & \yes
\\
\hline
\codebox{C} & 4 
& 
\texttt{Savor means to understand. Not to enjoy.}
& \no & \no & \textbf{?} & \noself
\\ 
\hline
\end{tabular}
}
\caption{
\textit{id: }116176c}
\vspace*{0.2cm}
\label{subtab:appx_annotation_varierr_704}
\end{subtable}
\begin{subtable}[h!]{1.0\textwidth}
\resizebox{\textwidth}{!}{
\begin{tabular}{ccl|cccc}
\multicolumn{7}{l}{\begin{tabular}[c]{@{}l@{}}
\textit{Premise}: 
\texttt{The tree-lined avenue extends less than three blocks to the sea.} \\
\textit{Hypothesis}: 
\texttt{The sea isn't even three blocks away.} \\
\textit{Label-explanation pairs:}
\texttt{before validation: \{``E'':4,``N'':1,``C'':1\}} \;
\texttt{Self-validated: \{``E'':3,``N'':1\}}
 \;
\texttt{Peer-validated: \{``E'':4,``N'':1\}} \\
\textit{Labels:} [E, N]
\quad
\textit{Error:} [\codebox{C}]
\end{tabular}} \\
\hline
\multicolumn{3}{c|}{\textit{Round 1: NLI Label \& Explanation}}
& \multicolumn{4}{c}{\textit{Round 2: Validity}}  \\
\multirow{1}{*}{L}  
& \multirow{1}{*}{A} 
& \multicolumn{1}{c|}{Explanation}
& 1 & 2 & 3 & 4 \\
\hline
 \multirow{4}{*}{E} & 1 & 
\texttt{Both premise and hypothesis talk about less than three blocks.}
& \yesself & \yes & \yes & \no
\\
 & 2 & 
\texttt{If the avenue reaches the sea after less than three blocks, it cannot be further away.}\hspace{1cm}
& \yes & \yesself & \yes & \no
\\
& 3 & 
\texttt{The avenue is less than three blocks to the sea.}
& \yes & \yes & \yesself & \no
\\
 & 4  & 
\texttt{If the hypothesis means that the sea is less than three blocks away.}
& \textbf{?} & \yes & \yes & \noself
\\
\hline
N & 3  & 
\texttt{It is not given where is the location of the narrator.}
& \yes & \no & \yesself & \yes
\\
\hline
\codebox{C} & 4  & 
\texttt{If the hypothesis means that the sea is more than three blocks away.}
& \textbf{?} & \no & \textbf{?} & \noself
\\
\hline
\end{tabular}
}
\caption{
\textit{id: }80630e}
\vspace*{0.2cm}
\label{subtab:appx_sample_annotation_varierr_1277}
\end{subtable}

\begin{subtable}[h!]{1.0\textwidth}
\centering
\resizebox{\textwidth}{!}{
\begin{tabular}{ccl|cccc}
\multicolumn{7}{l}{\begin{tabular}[c]{@{}l@{}}
\textit{Premise}: 
\texttt{As he stepped across the threshold, Tommy brought } 
\texttt{the picture down with terrific force on his head.} \\
\textit{Hypothesis}: 
\texttt{Tommy hurt his head bringing the picture down.} \\
\textit{Label-explanation pairs:}
\texttt{ before validation: \{``E'':3,``N'':1,``C'':1\}} \;
\texttt{Self-validated: \{``E'':3,``N'':1\}}
 \;
\texttt{Peer-validated: \{``E'':3,``N'':1\}} \\
\textit{Labels:} [E, N] \quad
\textit{Error:} [\codebox{C}]
\\
\end{tabular}} \\
\hline
\multicolumn{3}{c|}{\textit{Round 1: NLI Label \& Explanation}}
& \multicolumn{4}{c}{\textit{Round 2: Validity}}  \\
\multirow{1}{*}{L}  
& \multirow{1}{*}{A} 
& \multicolumn{1}{c|}{Explanation}
& 1 & 2 & 3 & 4 \\
\hline
\multirow{3}{*}{E} & 1
& 
\texttt{the picture 
hit Tommy in the head}
& \yesself & \yes & \yes & \no
\\
& 2 & 
\texttt{a picture 
hit Tommy's head with terrific force}
& \yes & \yesself & \yes & \no
\\
 & 3 & 
\texttt{Tommy hurt his head with the picture}
& \yes & \yes & \yesself & \no
\\
\hline
N & 3
& 
\texttt{ambiguous if Tommy hurt himself or another guy} \hspace{8.5cm}
& \yes & \yes & \yesself & \no
\\
\hline
\codebox{C} & 4
& 
\texttt{Tommy is not hurt but rather bad strong emotion}
& \no & \no & \yes & \noself
\\
\hline
\end{tabular}
}
\caption{
\textit{id: }77893n}
\label{subtab:appx-sample_annotation_varierr_1241}
\end{subtable}
\caption{Additional sample annotations 
from \name{} NLI corpus. L: Label, A: Annotator; E: Entailment, N: Neutral, C: Contradiction; \fbox{\textcolor{magenta}{magenta}}: self-judgments, black: peer-judgments, \codebox{Err}: label error.}
\label{tab:appx_sample_annotation_varierr}\end{table*}

\end{document}